\documentclass[journal]{IEEEtran}
\usepackage{amsfonts}
\usepackage{graphicx}
\usepackage[T1, OT1]{fontenc}
\DeclareTextSymbolDefault{\dh}{T1}
\usepackage{hyperref}
\usepackage{cite}
\usepackage[english]{babel}
\usepackage[utf8]{inputenc}
\usepackage[linesnumbered,ruled]{algorithm2e}
\usepackage[noend]{algpseudocode}
\usepackage{amsmath}
\usepackage{float}
\usepackage{caption}
\usepackage{color}
\usepackage{threeparttable}
\usepackage{multirow}
\ifCLASSINFOpdf
\else
\fi
\begin{document}
\title{Unsupervised Deformable Medical Image Registration via Pyramidal Residual Deformation Fields Estimation}
\author{Yujia Zhou, Shumao Pang, Jun Cheng, Yuhang Sun, Yi Wu, Lei Zhao, Yaqin Liu, Zhentai Lu, Wei Yang, and Qianjin Feng, \IEEEmembership{Member, IEEE}
\thanks{Qianjin Feng is the corresponding author.}
\thanks{This work was supported in part by the National Natural Science Foundation of China (No. 81801780, No. U1501256, and No. 81974275), and the Science and Technology Project of Guangzhou (No. 201704020033).}
\thanks{Y. Zhou, S. Pang, Y. Sun, Y. Wu, L. Zhao, Y. Liu, Z. Lu, and W. Yang are with Guangdong Provincial Key Laboratory of Medical Image Processing, School of Biomedical Engineering, Southern Medical University, Guangzhou, 510515, China. (e-mail: zyj.shmily08@gmail.com; pangshumao@126.com; 1821305331@qq.com; you1437@qq.com; lei6730@smu.edu.cn; liuyq@fimmu.com;  luzhentai@163.com; weiyanggm@gmail.com).}
\thanks{J. Cheng is with National-Regional Key Technology Engineering Laboratory for Medical Ultrasound, Guangdong Key Laboratory for Biomedical Measurements and Ultrasound Imaging, School of Biomedical Engineering, Health Science Center, Shenzhen University, Shenzhen, 518061, China. (e-mail: chengjun583@qq.com).}
\thanks{Q. Feng is with Guangdong Provincial Key Laboratory of Medical Image Processing, School of Biomedical Engineering, Southern Medical University, Guangzhou, 510515, China. (e-mail: fengqj99@fimmu.com)}}

\maketitle

\begin{abstract}
Deformation field estimation is an important and challenging issue in many medical image registration applications. In recent years, deep learning technique has become a promising approach for simplifying registration problems, and has been gradually applied to medical image registration. However, most existing deep learning registrations do not consider the problem that when the receptive field cannot cover the corresponding features in the moving image and the fixed image, it cannot output accurate displacement values. In fact, due to the limitation of the receptive field, the $3 \times 3$ kernel has difficulty in covering the corresponding features at high/original resolution. Multi-resolution and multi-convolution techniques can improve but fail to avoid this problem. In this study, we constructed pyramidal feature sets on moving and fixed images and used the warped moving and fixed features to estimate their "residual" deformation field at each scale, called the \textbf{Pyramidal Residual Deformation Field Estimation module (PRDFE-Module)}. The "total" deformation field at each scale was computed by upsampling and weighted summing all the "residual" deformation fields at all its previous scales, which can effectively and accurately transfer the deformation fields from low resolution to high resolution and is used for warping the moving features at each scale. Simulation and real brain data results show that our method improves the accuracy of the registration and the rationality of the deformation field.
\end{abstract}


\section{Introduction}
\label{sec:introduction}
\IEEEPARstart{D}{eformable} registration is a fundamental task in various medical imaging studies and has been a topic of active research for decades. In deformable registration, a dense, nonlinear correspondence is established between a pair of images, such as 3D magnetic resonance (MR) brain scans. Typically, deformable registration is solved by numerical optimization for each volume pair by aligning voxels with similar appearance such as mean square error (MSE) and normalized mutual information (NMI) while enforcing constraints on the registration mapping \cite{b1} -\cite{b3}. Unfortunately, solving a pairwise optimization can be computationally intensive and therefore slow in practice. 

To address the potential shortcomings of manual registration, deep learning techniques are being increasingly used for medical image registration \cite{b4} -\cite{b20}. Although supervised deep learning may still rely on ground truth voxel mappings, which has to be predicted from existing registration algorithms \cite{b4} -\cite{b6}, synthetic methods \cite{b7} -\cite{b9} or generative adversarial networks \cite{b10}, deep learning has changed the landscape of image registration research. Ever since the success of spatial transformer network (STN \cite{b11}), unsupervised deep learning techniques have allowed for state-of-the-art performance in many registration tasks. For example, unsupervised registration networks make a straightforward prediction for the deformation field between intensity images \cite{b12} or the corresponding features \cite{b13} based on the U-net structure. Weakly supervised methods attempt to learn the correspondences between more global or structural information of the images rather than that for voxel-level information, such as aligning the contours of the myocardium \cite{b14} and the segmentation of the prostate \cite{b15}.  Furthermore, some efforts have been made on the smoothness and rationality of the network output deformation field. Some of them introduce some constraints, such as the Inverse-Consistent \cite{b16} and Cycle-Consistent constraints \cite{b17}. Some learn a latent space of transformation parameters with low-rank \cite{b18} or diffeomorphic properties \cite{b19} \cite{b20} to guarantee the reasonability for the large displacements. 

The accuracy of registration depends on its ability to find the corresponding points in moving and fixed images regardless of the type of deep learning registration methods. However, all the above methods ignore a key issue: a correct displacement can be output only when the receptive field of the convolution kernel can cover the corresponding points in moving and fixed images (Fig. 1). Intuitively, if an image is $ \times 16$ downsampled, a kernel with a size of $3 \times 3$ can cover a size of $48 \times 48$ receptive field and easily find the corresponding points for most registration cases. However, a $3 \times 3$ kernel may not be able to cover the corresponding structures/features that are far away in the moving and fixed image when aligning an image at the original resolution. Consequently, no information can be conveyed for learning and output accurate deformation fields, and the gradient descent will remain in a poor local minimum. Typically, multiresolution and multiple convolutions can expand the receptive field, which can alleviate but not avoid the above problem, especially in the case of high resolution.

\begin{figure}[!h]
\centering
\includegraphics[width=2.8in]{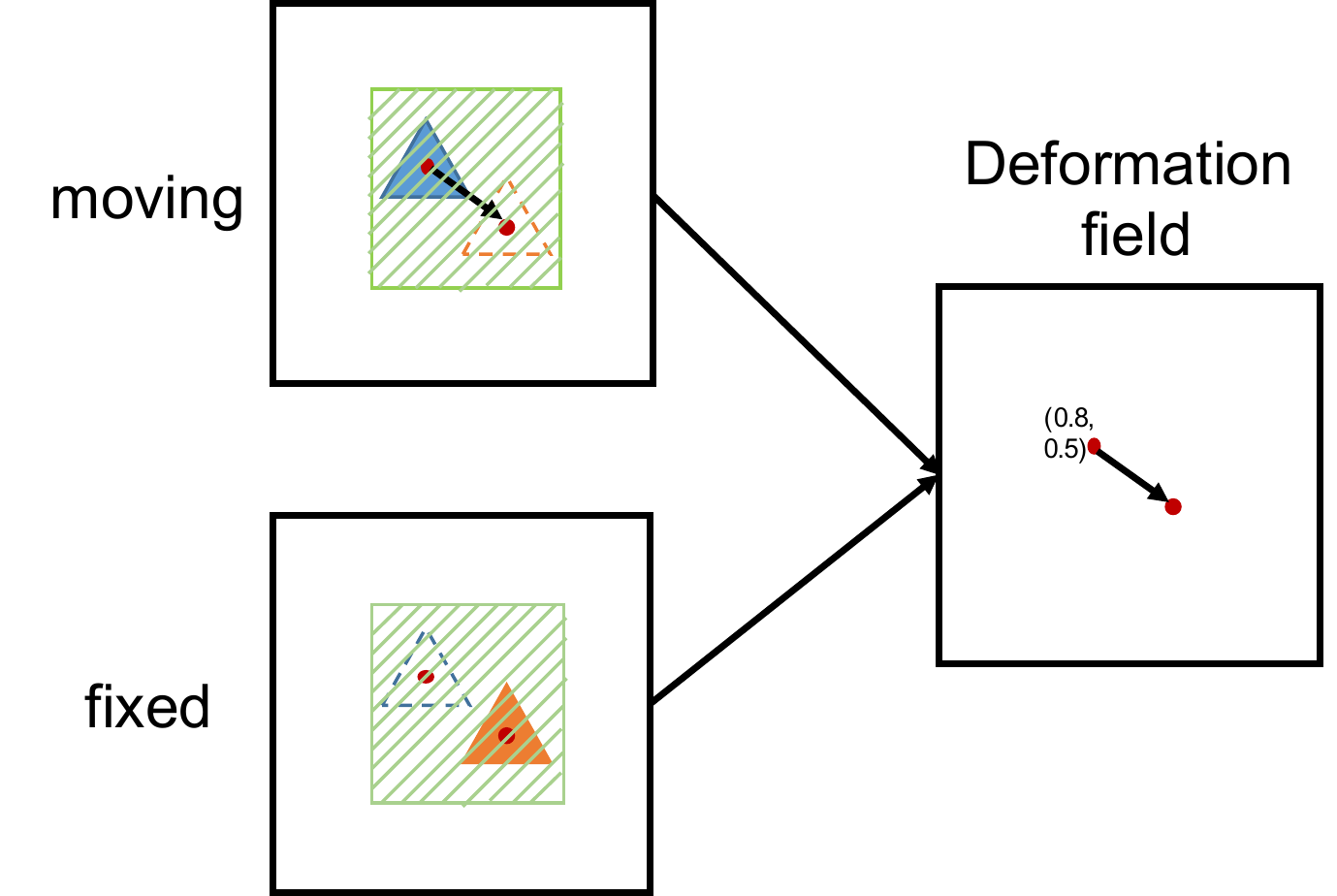}
\caption{Prerequisites for outputting the correct deformation field. Blue and Orange triangle are the corresponding structures in moving and fixed images.}
\label{fig_sim}
\end{figure}

In the past two years, the idea of warping first appeared in optical flow estimation \cite{b21} -\cite{b25}. The effectiveness of the idea of warping was proved from the beginning of image warping \cite{b21} \cite{b22} to the subsequent feature warping \cite{b23} -\cite{b25}. In the context of optical flow estimation, voxels in the images are highly correlated with each other both spatially and temporally. Conversely, in the case of image registration, the relationships between the corresponding voxels/structures are more complex without intuitively spatial or temporal regularity, which greatly increases the difficulty of accurately estimating the deformation field. A pioneered work \cite{b26} was proposed in this area, which estimated the total deformation field on each scale from the warped moving features and fixed features. However, two issues need to be further explained: first is when estimating the total deformation field on high resolution, the problem that the receptive field cannot cover the corresponding features still exists. In addition, all the above feature/image warping methods directly upsample the low-resolution deformation field to warp its subsequent high-resolution features but do not consider the problem that the amplitude of deformation fields needs to be also doubled in addition to its size. Therefore, it is difficult to accurately transfer the deformation field estimated at a low resolution to its subsequent high-resolution features. For example, if the value of the deformation field at a low-resolution point is 2, the value of the deformation field at that point should be doubled to 4 instead of 2 after upsampling. Otherwise, the points at the low resolution cannot be deformed to the correct specified position after the upsampling process, which greatly reduces the accuracy of the deformation field estimation.

Some cascaded/stacked networks \cite{b27} -\cite{b30} have been proposed to solve the above problem of the receptive field of convolution kernel in another way. They manually construct a series of images $\left \{ I_{1},..., I_{n}\right \}$ to describe the complex deformation process of moving and fixed images step by step. Then $n-1$ are cascaded to progressively learn the complex transformation relationships between moving and fixed images. The $k_{th}$ CNN with the inputs of $I_{k}$ and $I_{k+1}$ only needs to learn a limited distance of the corresponding structures in moving and fixed images. For example, Deep Learning Image Registration (DLIR) \cite{b27} cascades affine and nonlinear networks and trains them one by one, that is, after fixing the weights of previous cascades. Volume Tweening Network (VTN) \cite{b28} jointly trains the cascades, whereas all successively warped images are measured by the similarity compared with the fixed image. Recursive Cascaded Networks \cite{b29} enlarge the number of cascades and only measure the similarity of the final cascade. A morphological simplification network (MS-Net) \cite{b30} is another type of cascaded network, which gradually warps cortex-folding patterns rather than images. In conclusion, the above deep learning registration methods are devoted to simplify the large deformation on a carefully handcrafted space and suffer from the following limitation: inaccurate deformation field estimation at a cascade/stack will lead to artifacts in the warped moving image/features and be propagated into the following cascade/stack registration. Each cascade/stack input, the time for early stopping, and the weights between each cascade/stack loss require to be carefully handcrafted.

\begin{figure*}[!h]
      \centering
      \includegraphics[width=6.5in]{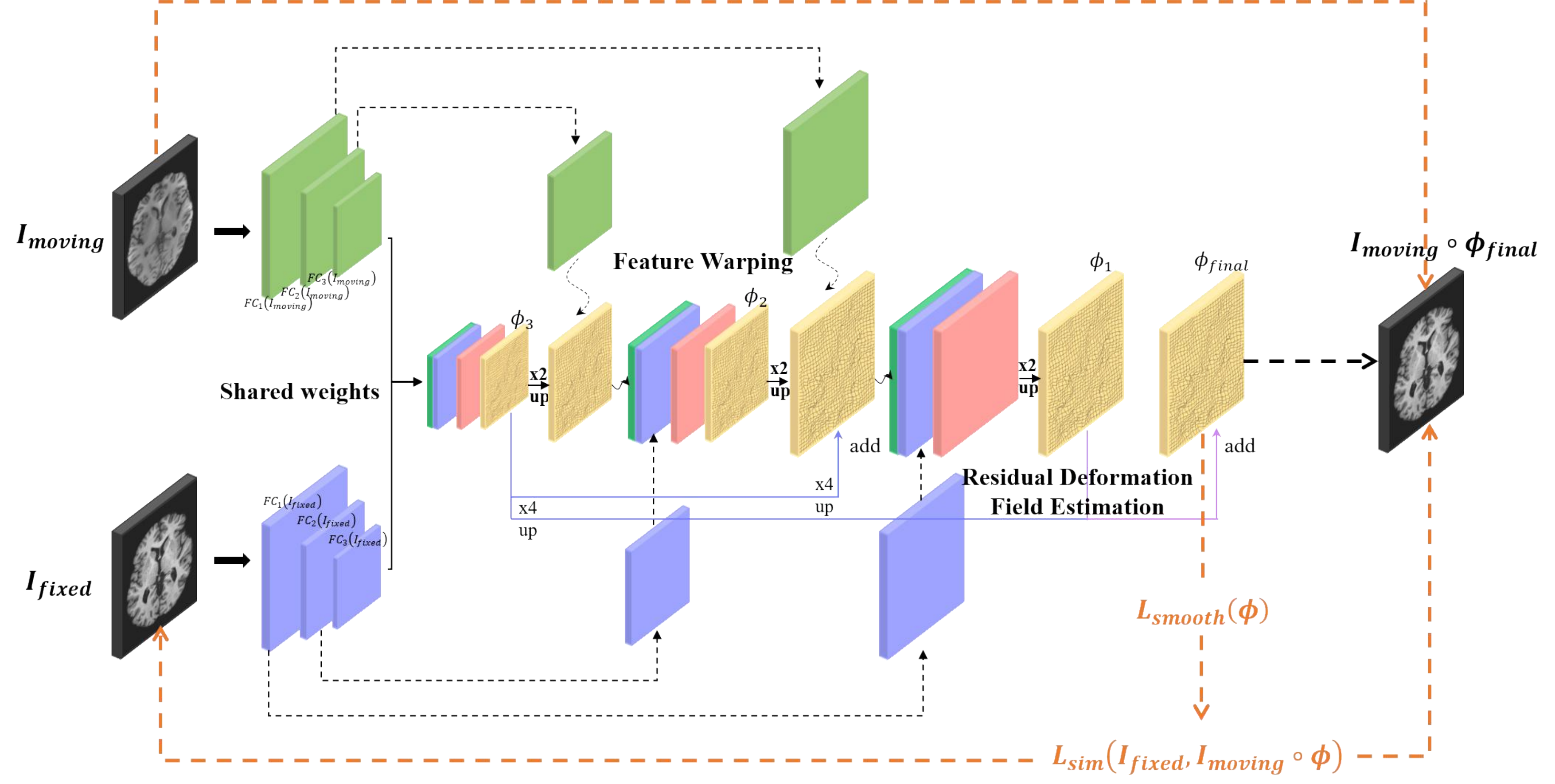}
      \caption{The overview of our proposed PRDFE-Module on the basis of U-net (3 layers for example).}
      \label{fig_sim}
\end{figure*} 

In the present study, we proposed a simple but effective module with clear interpretability, called a \textbf{Pyramidal Residual Deformation Fields Estimation Module (PRDFE-Module)}. First, instead of an image pyramid, we train a pyramidal feature descriptor that resembles image similarity for establishing a robust correspondence. The image pair is transformed from the spatial domain to a learnable feature space in forms of two pyramids of multiscale feature maps. Second, a feature warping module is established at each scale of the moving feature pyramid to gradually warp the moving features to the fixed features. Third, a residual deformation estimation and propagation module computes two deformation fields involving in the feature warping module: one is the residual deformation field; the other is the total deformation field. The former comes from convolutional output of the concatenated fixed and warped moving features. The latter, which is used for warping the moving features of the next scale, comes from the weighted summation from residual deformation fields of all previous scales after resampling. Finally, when the deformation field has the same resolution as the input image, it serves as the output deformation field of the network to warp the moving image. In this way, the calculated loss function between the warped image and the fixed image is used to train the entire network. The benefits of our proposed module are obvious:

\begin{itemize}
\item \textbf{Facilitates the corresponding feature finding in the receptive field}: Regardless of feature warping or residual deformation field estimation, the final goal is to make the corresponding features in moving and fixed images easy to find in a limited size of receptive field on each scale. The estimation of the residual deformation field can split large and complex deformation into a series of small and local ones, ensuring that the displacements of warped moving and fixed features are small and within the range of the receptive field, especially for high-resolution estimation.  
\item \textbf{Effectively transfers low-resolution deformation fields to high-resolution ones}: In our module, the output/total deformation field at each scale is the weighted sum of residual deformation fields at all previous scales. In other words, we enlarge not only the size of the residual deformation field but also its magnitude value. Such weighted summation ensures that the residual deformation fields can warp features to the correct positions after upsampling and ensures the accuracy of the deformation field when it is transferred from low resolution to high resolution.

\end{itemize}

\section{Method}
The goal of image registration is to determine a deformation field $\phi : \mathbb{R}^{3} \rightarrow \mathbb{R}^{3}$ that warps a moving image ${I_{moving}} \in \mathbb{R}^{3}$ to a fixed image ${I_{fixed}} \in \mathbb{R}^{3}$. Fig.2 presents an overview of our method. On the basis of U-net, Shared-Weight Pyramidal Feature Descriptor, Feature Warping Module, and Residual Deformation Field Estimation and Propagation Module are embedded. \textbf{Shared-Weight Pyramidal Feature Descriptor} encodes ${I_{moving}}$ and ${I_{fixed}}$ respectively into two pyramids of multi-scale features $\left \{ {FE}_{k} \left ( I_{moving}, k=1,...,K \right ) \right \}$ and $\left \{ {FE}_{k} \left ( I_{fixed}, k=1,...,K \right ) \right \}$; \textbf{Feature Warping Module} warps each scale of features on the pyramid ${FE}_{k} \left ( I_{moving} \right )$ with the deformation field $\phi _{k}$; \textbf{Residual Deformation Field Estimation Module} estimates and propagates the residual deformation field from the warped moving and fixed features. The above modules can be directly applied into an unsupervised network, such as VoxelMorph \cite{b12}, with the intensity images as inputs or a weakly-supervised network, such as the network in \cite{b15} with the intensity images and segmentations as training inputs.

\subsection{Shared-Weight Pyramidal Feature Descriptor}
Fig. 2 shows that encoding is a two-stream subnetwork, in which the filter weights are shared across the two streams. Each of them functions as a pyramidal feature descriptor that transforms an image  to a pyramid of multiscale features ${FE}_{k} \left ( I \right )$ from the highest spatial resolution $(k=1)$ to the lowest spatial resolution $(k=K)$. To generate feature representation at the $k_{th}$ scale ${FE}_{k} \left ( I \right )$, we use cascaded convolution and pooling layers by a downsample factor of 2 at the $(k-1)_{th}$ pyramid level. The Pyramidal Feature Descriptor can be enhanced with ResNet or DenseNet connections. 

\subsection{Feature Warping Module}
Deformation field inference becomes more challenging if ${I_{moving}}$ and ${I_{fixed}}$ are captured far away from each other because a correspondence can only be found in the local receptive fields of a convolution layer. Inspired by the use of image warping to address large deformation in conventional registration methods, we proposed to reduce feature–space distance between $\left \{ {FE}_{k} \left ( I_{moving}, k=1,...,K \right ) \right \}$ and $\left \{ {FE}_{k} \left ( I_{fixed}, k=1,...,K \right ) \right \}$ through the Feature Warping layer. $\left \{ {FE}_{k} \left ( I_{moving}, k=1,...,K \right ) \right \}$ is warped toward $\left \{ {FE}_{k} \left ( I_{fixed}, k=1,...,K \right ) \right \}$ by STN via the total deformation field $\phi _{k}$ at the $k_{th}$ pyramid level:
\begin{equation}
\tilde{{FE}_{k}} \left ( I_{moving} \right )  = \phi _{k} \circ {FE}_{k} \left ( I_{moving} \right )
\end{equation}

The estimation of $\phi _{k}$ is introduced at Section 2.3.

\subsection{Deformation Field Estimation and Propagation}

First, a residual deformation field $\phi _{residual}^{k}$ at the $k$-level pyramid is computed:
\begin{equation}
\phi _{residual}^{k} = G_{k}\left ( {FE}_{k}\left ( I_{fixed} \right ), \tilde{{FE}_{k}}\left ( I_{moving} \right ) \right )
\end{equation}

where $G_{k}(.)$ is a convolutional block (or a ResNet/DenseNet block) at the $k_{th}$ pyramid level.

However, the residual deformation field $\phi _{residual}^{k}$ at the $k_{th}$ scale cannot be directly upsampled and applied to ${FE}_{k+1} \left ( I_{moving} \right )$. There are two issues to consider: First, as no warping has been processed on ${FE}_{k+1} \left ( I_{moving} \right )$, we have to sum the residual deformation fields on its all previous scales $\left \{ 1,...,k \right \}$. Second, we need to weigh the deformation fields of different resolutions. For example, the magnitude of $down_2$ deformation field should be doubled, and the magnitude of $down_4$ deformation field should be multiplied by 4, and so on. By applying this method, the low-resolution deformation fields can guide each voxel on the moving features to the corresponding position during the upsampling process. 

Therefore, the total deformation $\phi _{residual}^{k}$ at the $k$-level pyramid needs to combine all residual deformation fields $\left \{ \phi _{residual}^{j}, j=1,...,k-1  \right \}$ in a balanced way, that is, resampling and weighted summing operations:
\begin{equation}
\phi _{k} = \sum_{j=1}^{k} 2^{j-1} \times up_{2_{j-1}} (\phi _{residual}^{k})
\end{equation}

\begin{figure*}[!h]
      \centering
      \includegraphics[width=7in]{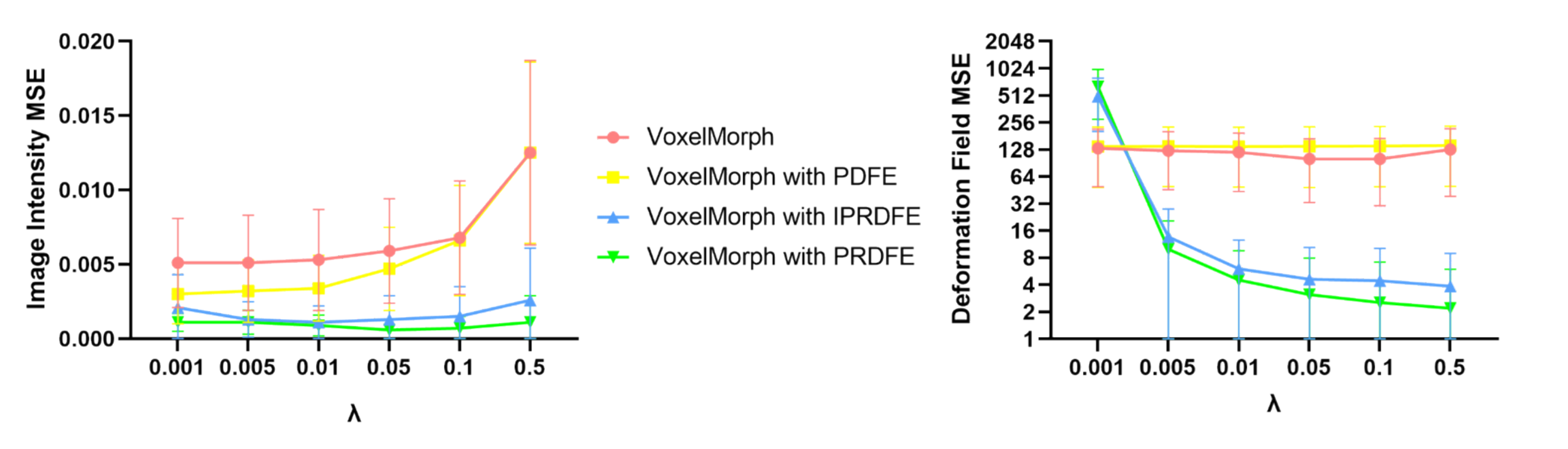}
      \caption{Image Intensity MSE and Deformation Field MSE with respect to ground truth with different regularization parameters.}
      \label{fig_sim}
\end{figure*} 

\subsection{Training Loss}
Given that the multi-resolution loss calculation may disturb the gradient optimization and decrease the convergence speed, we only measure the similarity on the original resolution warped image, enabling all scales to learn progressive alignments cooperatively.

Specifically, deformation fields at different scales have been upsampled to the same size, that is, the highest spatial resolution $(k=1)$. Moreover, the final output deformation field $\phi$ that warps the moving image to the fixed image is obtained by weighing the residual deformation fields of all previous scales.

\begin{multline}
\phi _{final} = \phi = 16\times up_{16} \phi _{residual} ^ {5} + 8\times up_{8} \phi _{residual} ^ {4} \\
+ 4\times up_{4} \phi _{residual} ^ {3} +2\times up_{2} \phi _{residual} ^ {2} + 1\times \phi _{residual} ^ {1} 
\end{multline}

Then, the training loss can be defined as the weighted sum of the MSE between the warped image and the moving image and the $l_2-$norm gradients \cite{b33} of the deformation field $\phi _{final}$

\begin{multline}
L_{intensity} = MSE(I_{fixed}, \phi _{final} \circ I_{moving}) + \lambda Grad( \phi _{final}) \\
= \frac{1}{\Omega }\sum_{p\in \Omega } \left [ I_{fixed}(p)-\left [  \phi _{final} \circ I_{moving} \right ](p) \right ]^{2} \\
+ \lambda \sum_{p\in \Omega }(\left \| \frac{\partial \phi _{final}(p) }{\partial x} \right \|^{2} + \left \| \frac{\partial \phi _{final}(p) }{\partial y} \right \|^{2} + \left \| \frac{\partial \phi _{final}(p) }{\partial z} \right \|^{2})
\end{multline}

in which $p$ is a voxel and $\lambda$ is a regularization parameter.

\section{Materials and Assessment}

\subsection{Dataset}
\subsubsection{Simulated Data}
We constructed 10 types of geometric figures with textures and generated a series of affine + nonlinear deformation fields through the SyN \cite{b1} toolkit. Finally, 2500 2D images were generated. Among them, 2000 and 500 images were used for training and testing, respectively. The size of the above images was 128 x 128 and the intensity range is [0, 1]. No preprocessing scheme was used. 

\subsubsection{Real Brain Data}

We evaluated our method on a large-scale, multisite, multistudy dataset of 2953 T1-weighted 3D brain MR images. Eight public datasets, namely, OASIS \cite{b32}, ADHD \cite{b33}, ABIDE \cite{b34}, PPMI \cite{b35}, LPBA40 \cite{b36}, ISBR18 \cite{b37}, CUMC12 \cite{b37} and MGH10 \cite{b37}, were used. We split our dataset into 2623 and 250 volumes for the training and validation sets, respectively. In addition, 80 volumes in the LPBA40 (40 brain images, each with 56 manually-labeled ROIs), ISBR18 (18 brain images, each with 96 manually-labeled ROIs), CUMC12 (12 brain images, each with 130 manually-labeled ROIs) and MGH10 (10 brain images, each with 106 manually-labeled ROIs) were only used for testing. Standard preprocessing steps, including affine spatial normalization, histogram matching, and rescaling to [0,1], were performed, and the resulting images were cropped to 160 x 192 x 224. For the training data, we resized the input images to 120 x 144 x 160 due to limited graphics processing unit (GPU) memory. For the testing data, we used the images with the size of 160 x 192 x 224 considering the processing steps in VoxelMorph \cite{b12}.

\begin{figure*}[!h]
      \centering
      \includegraphics[width=7in]{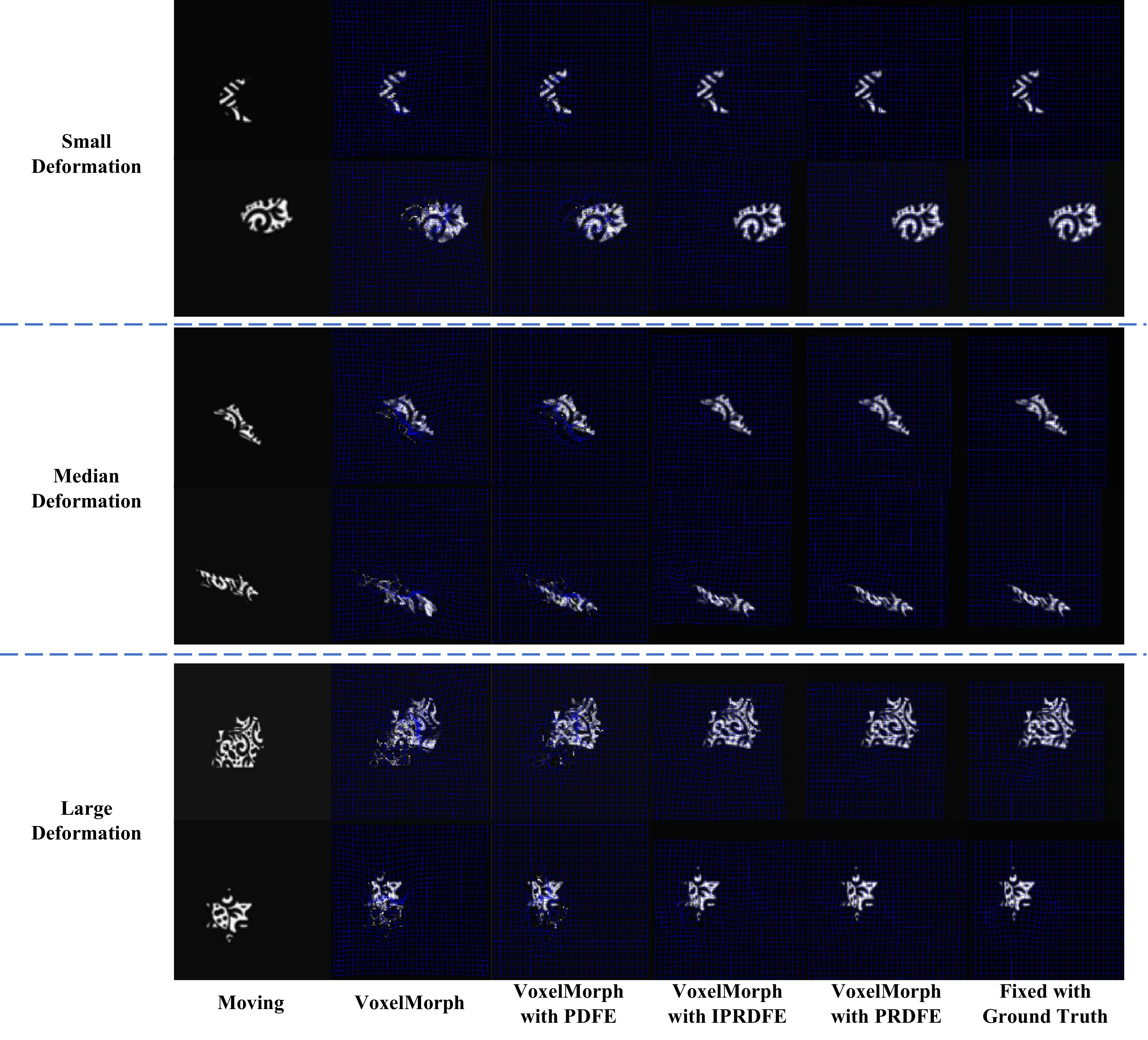}
      \caption{Visualization registration performance of 6 examples with Large, Median and Small Deformation in simulated dataset. Green box indicates inaccurate alignment results and orange box indicates unreasonable deformable registration.}
      \label{fig_sim}
\end{figure*} 

\subsection{Evaluation Metrics}

\subsubsection{Simulated Data}
As the simulated data were aligned by a designed deformation field, a ground truth can be used to evaluate registration accuracy. A known transform $\phi _{g} \left (\mathbf{p}  \right )$ was applied to the moving data and deformed to the fixed data using different deep learning registration methods. Therefore, the accuracy of the registration method can be evaluated by calculating MSE between (1) the warped image output from different networks and the known fixed images (small values are better) and (2) the obtained transforms using different registration schemes $\phi \left (\mathbf{p}  \right )$ and the known transform  $\phi _{g} \left (\mathbf{p}  \right )$ (small values are better). 
\subsubsection{Real Brain Data}
No ground truth was available for the real brain data. We adopted Dice similarity coefficient (DSC) of the annotated tissues  to quantitatively evaluate the registration performance of real brain data.

\begin{figure*}[!h]
      \centering
      \includegraphics[width=6.5in]{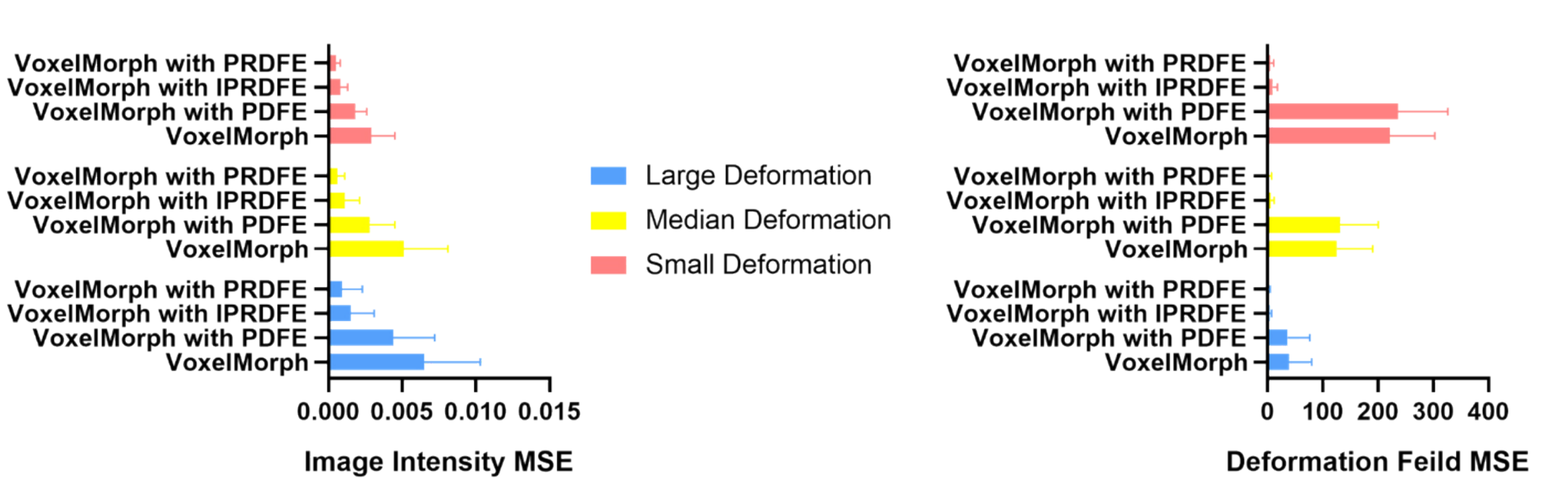}
      \caption{Image Intensity MSE and Deformation Field MSE with respect to Large, Median and Small Deformation.}
      \label{fig_sim}
\end{figure*}

\begin{table*} \label{table number}
\centering
\caption{Statistical test results on comparisons between different registration methods}
\begin{threeparttable}
\begin{tabular}{ccccc}
\hline
& VoxelMorph & VoxelMorph with PDFE & VoxelMorph with IPRDFE & VoxelMorph with PRDFE \\
\hline
Image Intensity MSE & $0.0051\pm0.0030$ & $0.0030\pm0.0020$ & $0.0011\pm0.0011$ & $\mathbf{0.0006\pm0.0007}$ \\
\hline
Deformation Field MSE & $132.2309\pm83.0509$ & $138.4297\pm90.0017$ & $6.0331\pm6.5563$ & $\mathbf{3.1005\pm4.8268}$ \\
\hline
\end{tabular}
\end{threeparttable}
\end{table*}
\subsection{Comparison Methods}

We select three state-of-the-art registration methods, including two traditional methods (Diffeomorphic Demons (\textbf{D.Demons}) \cite{b39} and \textbf{SyN} \cite{b1}), and one deep learning method (\textbf{VoxelMorph} \cite{b12}), for comparison. Furthermore, we performed experiments with/without some components of our PRDFE-Module, without pyramidal feature and residual warping components, and marked them as \textbf{VoxelMorph with IPRDFE} (Image Pyramidal Residual Deformation Fields Estimation), \textbf{VoxelMorph with PDFE} (pyramidal deformation field estimation without “residual” module), and \textbf{VoxelMorph with PRDFE} (our proposed module). 

\subsection{Implementation details}

We implemented our method based on the CVPR-version code \footnote[1]{https://github.com/voxelmorph/voxelmorph} in VoxelMorph-2 with MSE loss. Each network was trained with a 12GB NVIDIA® Pascal™ TITAN Xp general-purpose GPU on a high-performance computing cluster for 1 h on simulated data and 48 h on real brain data, separately. All parameters were set as default, including the ADAM optimizer, a learning rate of 1e-4, and a mini-batch of 4. For the simulated data, the regularization parameter with the highest image registration accuracy was selected. Moreover, the number for each encoding feature channel is [144, 160, 160, 160] and [160, 160, 160, 160, 160, 144, 144] for each decoding feature channel. For brain data, all registration methods chose a regularization of 0.001. The number for each encoding feature channel is [24, 40, 40, 40] and [40, 40, 40, 40, 40, 24, 24] for each decoding feature channel.

\section{Experimental Results}

\begin{table*} \label{table number}
\centering
\caption{Results for IBSR18, CUMC12, MGH10, LPBA40 in term of DSC (\%).}
\begin{threeparttable}
\begin{tabular}{cccccccc}
\hline
\multirow{2}{*}{\begin{tabular}[c]{@{}c@{}} Dataset \\ \end{tabular}} &
\multirow{2}{*}{\begin{tabular}[c]{@{}c@{}} Brain Tissue \\ \end{tabular}} &
\multirow{2}{*}{\begin{tabular}[c]{@{}c@{}} D.Demons \\ \end{tabular}} &
\multirow{2}{*}{\begin{tabular}[c]{@{}c@{}} SyN \\ \end{tabular}} &
\multirow{2}{*}{\begin{tabular}[c]{@{}c@{}} VoxelMorph \\ \end{tabular}} &
\multirow{2}{*}{\begin{tabular}[c]{@{}c@{}} VoxelMorph \\ with PDFE \end{tabular}} &
\multirow{2}{*}{\begin{tabular}[c]{@{}c@{}} VoxelMorph \\ with IPRDFE \end{tabular}} &
\multirow{2}{*}{\begin{tabular}[c]{@{}c@{}} VoxelMorph \\ with PRDFE \end{tabular}} \\
\\
\hline
ISBR18 & GM & $84.79 \pm 2.36$ & $83.40 \pm 2.24$ &  $84.10 \pm 5.53$ & $83.98 \pm 4.01$ & $85.14 \pm 3.82$ & $\mathbf{85.37\pm 4.55}$ \\

& WM & $83.41 \pm 3.40$ & $80.38 \pm 2.32$ & $81.39 \pm 6.27$ & $81.20 \pm 4.83$ & $82.77 \pm 4.21$ & $\mathbf{83.94 \pm 4.27}$ \\

CUMC12 & GM & $76.81 \pm 4.61$ & $75.38 \pm 7.23$ & $77.46 \pm 6.61$ & $77.53 \pm 5.74$ & $78.91 \pm 4.41$ & $\mathbf{80.00 \pm 4.68}$ \\

& WM & $85.11 \pm 3.12$ & $82.98 \pm 4.65$ & $84.52 \pm 4.40$ & $84.53 \pm 3.90$ & $86.12 \pm 2.85$ & $\mathbf{87.46 \pm 2.88}$ \\

MGH10 & GM & $82.35 \pm 4.17$ & $80.94 \pm 6.17$ & $83.22 \pm 5.24$ & $83.39 \pm 4.59$ & $84.29 \pm 3.8$ & $\mathbf{84.93 \pm 3.53}$ \\

& WM & $85.86 \pm 3.98$ & $84.39 \pm 4.89$ & $86.57 \pm 4.09$ & $86.60 \pm 3.64$ & $87.71 \pm 2.9$ & $\mathbf{88.74 \pm 2.69}$ \\

LPBA40 & GM & $80.19 \pm 5.34$ & $77.17 \pm 5.87$ & $82.55 \pm 2.93$ & $83.06 \pm 2.67$ & $83.61 \pm 2.42$ & $\mathbf{85.18 \pm 2.32}$ \\

& WM & $83.03 \pm 3.89$ & $80.41 \pm 4.26$ & $85.78 \pm 2.34$ & $85.92 \pm 2.20$ & $86.86 \pm 1.92$ & $\mathbf{88.63 \pm 1.88}$ \\
\hline
\end{tabular}
\end{threeparttable}
\end{table*}

\subsection{Simulated Data}

\subsubsection{Sensitivity to Regularization Parameter}

We conducted a simulation-based experiment to investigate the sensitivity of different deep learning registration methods to the involved regularization parameters (i.e., $\lambda$). We directly computed the MSE between warped and fixed image and that of output deformation field with ground truth deformation field under the condition of different regularization parameters (Fig. 3). The image intensity MSE of all deep learning methods was lower when the weight of the regularization term was small. This may be due to the high degree of freedom of the deformation field, which causes a point to be warped to multiple positions, and the interpolation results in an image that is highly similar to the fixed image. As the weight of the regularization term gradually increased, the image intensity MSE of VoxelMorph and VoxelMorph with PDFE increased sharply, indicating that once there is a tendency to one-to-one alignment, its registration accuracy will be severely reduced. However, the image intensity MSE of VoxelMorph with IPRDFE and VoxelMorph with PRDFE still maintained a low level, and the registration accuracy was stable. Moreover, the deformation field MSE of VoxelMorph with IPRDFE and VoxelMorph with PRDFE also maintained a low level, except at $\lambda = 0.001$. Finally, image MSE showed the best performance separately in VoxelMorph, VoxelMorph with PDFE, VoxelMorph with IPRDFE, and VoxelMorph with PRDFE when $\lambda$ = 0.001, 0.001, 0.01, and 0.05, respectively. Therefore, the above parameters were chosen to evaluate the registration performance of these methods.

\begin{figure*}[!h]
      \centering
      \includegraphics[width=7in]{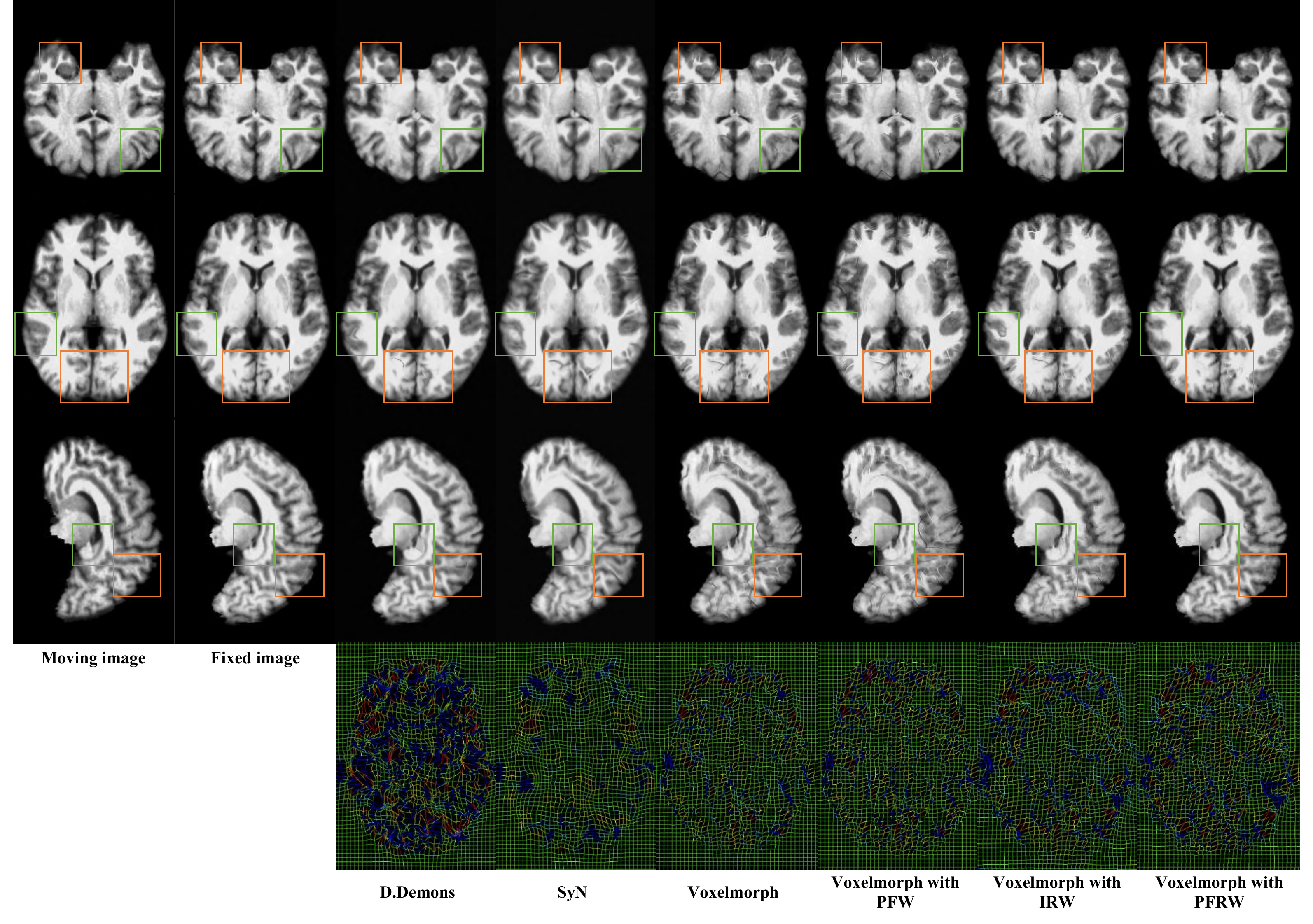}
      \caption{Visualization registration performance of an example in LPBA40 dataset. Green box indicates inaccurate alignment results and orange box indicates unreasonable deformable registration.}
      \label{fig_sim}
\end{figure*} 
\subsubsection{Registration Performance with Different Deep Learning Registration methods}

Figs. 4 and 5 show the visualization and quantitative results of different deep learning registration methods on simulated data. Although VoxelMorph with PDFE is superior to VoxelMorph in evaluating the MSE of warped and fixed images, the deformation output from VoxelMorph with PDFE is still unreasonable. This result is because neither VoxelMorph nor VoxelMorph with PDFE considers whether the receptive field of the convolution kernels covers the corresponding features in moving and fixed images. When considering the above problem, VoxelMorph with IPRDFE and VoxelMorph with PRDFE both achieve good registration accuracy and reasonable deformation fields. However, our method (VoxelMorph with PRDFE) has the best registration accuracy and the most reasonable deformation field, especially for large deformations. All quantitative results are listed in Table I.

\subsection{Real Brain Data}
We performed registrations for a total of $N \times (N-1)$ possible pairs of template and subject images for each registration method under comparison, where $N$ is the number of images in the testing dataset. The evaluation results are summarized in Table II. The mean and standard deviation of DSC were provided. In the four testing datasets, the DSC scores of our proposed module (VoxelMorph with PRDFE) were higher than those of the basic VoxelMorph and other VoxelMorph with only some components. Moreover, the DSC scores for VoxelMorph with PRDFE achieved the best performance among all registration methods, including traditional registration and other deep learning registration methods. The visualization of the registration results in Fig. 6 shows that our proposed method achieved more accurate surface alignment especially in the regions indicated by the green box and more reasonable deformable results in the regions indicated by the orange box compared with other methods.

\section{Discussion}
The main concern in traditional registration and deep learning registration methods is to find the corresponding features in moving and fixed images. However, current deep learning registration methods do not consider an essential problem: the deep learning registration method can only search for corresponding features within the limited receptive field of the convolution kernel. If there are no corresponding features in the receptive field of the convolution kernel, deep learning networks will only push/pull the points in the receptive field randomly and obtain a warped image similar to the fixed image by interpolation. This is not a “true” registration process, and its output deformation field is not reasonable. Although multiple convolutions can slightly increase the receptive field of the convolution kernel and multiple downsampling can increase the receptive field of the convolution kernel at low resolution registration, this problem still exists in high-resolution registration.

Our proposed method solves this problem with simple but effective techniques. The first is the shared weight pyramidal feature descriptor. The moving and fixed images were convoluted with the same kernels, which capture the corresponding features of different scales existing in both moving and fixed images. Determining the correspondence for multiscale feature descriptors is obviously more robust than for the intensity of a voxel. The second is the feature warping module. A small searching range rather than a large searching range is used to establish voxel correspondences because the feature space distance between the images has been reduced by a feature warp layer at each pyramid level. This effectively reduces the computational burden raised by the explicit matching. The third is the residual deformation field estimation and weighted summation module. Although the residual deformation field reduces the difficulty of solving the large deformation field, directly applying the residual deformation field to the motion features on the next scale may interfere with the accurate output of the deformation field. The weighted summation module can balance the residual deformation fields of different scales, so that all scales can cooperate to learn progressive alignment. In this way, the corresponding features with large deformation are covered by the receptive field of multiple $3 \times 3$ convolutional kernels at the lower-resolution stage. Then, a rough estimation of displacements is output and carefully and accurately applied to the higher-resolution stage such that moving features are closer to their corresponding fixed features. The higher-resolution stage, which receives the original fixed and warped moving features as inputs, only needs to handle smaller residual deformations to further refine the deformation field estimation. 

The visualization and quantification experiments on simulation data have proven our concerns and the effectiveness of our proposed module. VoxelMorph and VoxelMorph with PDFE do not consider whether the receptive field of the convolution kernel covers the corresponding features in moving and fixed images and cannot achieve “real” registration. It can only rely on reducing the regularization term weight and enlarging the degree of the deformation field to achieve the warped image; that is, a point is moved to multiple positions, and an image that looks reasonable and similar to the fixed image is obtained through interpolation.

Moreover, our results on real brain data demonstrated that our proposed VoxelMorph with PRDFE improved registration accuracy through a better and more reasonable estimation of the large deformation field compared with other methods. Our proposed method could improve its registration performance (Table II for DSC of GM/WM brain structures) compared with basic VoxelMorph.

Some limitations exist in our work. First, several preprocessing steps, including affine registration and histogram matching, are needed in our framework. The former is used to simplify the optimization of subsequent and more complex deformable image registration. Although it can be alleviated by propagating the coarse-level displacements and feature warping module, it is better to jointly train a global and local subnetwork if there is significant memory usage. The latter is used to handle domain differences among different brain datasets. Without this, the time and complexity of training will increase greatly.

Second, compared with traditional registration method, the output deformation field of real brain data lacks rationality, in which more percentages of voxels with a non-positive Jacobian determinant exist. This limitation appears in all deep learning registration frameworks. How to ensure the rationality of the deformation field and improve the accuracy of registration is one of our future research directions.

\section{Conclusion}

In this study, we introduced a residual deformation field estimation network with pyramidal feature warping module, which ensures that the corresponding features of moving and fixed images are within the receptive field of the convolution kernel. The burden of the large deformation between images can be alleviated by iteratively reducing the distance of the feature space in a coarse-to-fine manner. Simulation results and real brain data show that the registration accuracy can be improved and that the deformation field is more reasonable when applying our proposed module on VoxelMorph. In summary, the proposed model could be directly applied to many practical registration problems because it is an accurate and easy-to-embed method for deep learning registration framework.


\begin{thebibliography}{1}

\bibliographystyle{unsrt}

\bibitem{b1} B. B. Avants, C. L. Epstein, M. Grossman, and J. C. Gee, ``Symmetric diffeomorphic image registration with cross-correlation: evaluating automated labeling of elderly and neurodegenerative brain,'' \emph{Medical image analysis}, vol. 12, no. 1, pp. 26-41, 2008.

\bibitem{b2} M. F. Beg, M. I. Miller, A. Trouvé, and L. Younes, ``Computing large deformation metric mappings via geodesic flows of diffeomorphisms,'' \emph{International journal of computer vision}, vol. 61, no. 2, pp. 139-157, 2005.

\bibitem{b3} R. Bajcsy and S. Kovačič, ``Multiresolution elastic matching,'' \emph{Computer vision, graphics, and image processing}, vol. 46, no. 1, pp. 1-21, 1989.

\bibitem{b4} X. Yang, R. Kwitt, M. Styner, and M. Niethammer, ``Quicksilver: Fast predictive image registration - A deep learning approach,'' \emph{Neuroimage}, vol. 158, pp. 378-396, Sep 2017.

\bibitem{b5} X. Cao, et al, ``Deformable image registration based on similarity-steered CNN regression,'' \emph{in International Conference on Medical Image Computing and Computer-Assisted Intervention}, 2017, pp. 300-308: Springer.

\bibitem{b6} K. A. Eppenhof, M. W. Lafarge, P. Moeskops, M. Veta, and J. P. Pluim, ``Deformable image registration using convolutional neural networks,'' \emph{in Medical Imaging 2018: Image Processing}, 2018, vol. 10574, p. 105740S: International Society for Optics and Photonics.

\bibitem{b7}  H. Uzunova, M. Wilms, H. Handels, and J. Ehrhardt, ``Training CNNs for Image Registration from Few Samples with Model-based Data Augmentation,'' \emph{in International Conference on Medical Image Computing \& Computer-assisted Intervention}, 2017.

\bibitem{b8}  H. Sokooti, B. de Vos, F. Berendsen, B. P. Lelieveldt, I. Išgum, and M. Staring, ``Nonrigid image registration using multi-scale 3D convolutional neural networks,'' \emph{in International Conference on Medical Image Computing and Computer-Assisted Intervention}, pp. 232-239: Springer, 2017.

\bibitem{b9}  J. Krebs, et al, ``Robust non-rigid registration through agent-based action learning,'' \emph{in International Conference on Medical Image Computing and Computer-Assisted Intervention}, pp. 344-352: Springer, 2017.

\bibitem{b10}   D. Mahapatra, B. Antony, S. Sedai, and R. Garnavi, ``Deformable medical image registration using generative adversarial networks,'' \emph{in 2018 IEEE 15th International Symposium on Biomedical Imaging (ISBI 2018)}, 2018.

\bibitem{b11}   M. Jaderberg, K. Simonyan, and A. Zisserman, ``Spatial transformer networks,'' \emph{in Advances in neural information processing systems}, pp. 2017-2025, 2015.

\bibitem{b12}   G. Balakrishnan, A. Zhao, M. R. Sabuncu, J. Guttag, and A. V. Dalca, ``VoxelMorph: a learning framework for deformable medical image registration,'' \emph{IEEE transactions on medical imaging}, 2019.

\bibitem{b13}   G. Wu, M. Kim, Q. Wang, B. C. Munsell, and D. Shen, ``Scalable high-performance image registration framework by unsupervised deep feature representations learning,'' \emph{IEEE Transactions on Biomedical Engineering}, vol. 63, no. 7, pp. 1505-1516, 2015.

\bibitem{b14}   M.-M. Rohé, M. Datar, T. Heimann, M. Sermesant, and X. Pennec, ``SVF-Net: Learning deformable image registration using shape matching,'' \emph{in International Conference on Medical Image Computing and Computer-Assisted Intervention}, pp. 266-274: Springer, 2017.

\bibitem{b15}   Y. Hu, et al, ``Weakly-supervised convolutional neural networks for multimodal image registration,'' \emph{Medical Image Analysis}, vol. 49, p. 1, 2018.

\bibitem{b16}   J. Zhang, ``Inverse-consistent deep networks for unsupervised deformable image registration,'' \emph{arXiv preprint arXiv:1809.03443}, 2018.
\bibitem{b17}   B. Kim, et al, ``Unsupervised Deformable Image Registration Using Cycle-Consistent CNN,'' \emph{arXiv preprint arXiv:1907.01319}, 2019.

\bibitem{b18}   J. Krebs, H. e Delingette, B. Mailhé, N. Ayache, and T. Mansi, ``Learning a probabilistic model for diffeomorphic registration,'' \emph{IEEE transactions on medical imaging}, 2019.

\bibitem{b19}   A. V. Dalca, G. Balakrishnan, J. Guttag, and M. R. Sabuncu, ``Unsupervised learning for fast probabilistic diffeomorphic registration,'' \emph{in International Conference on Medical Image Computing and Computer-Assisted Intervention}, pp. 729-738: Springer, 2018.

\bibitem{b20}   L. Liu, X. Hu, L. Zhu, and P.-A. Heng, ``Probabilistic Multilayer Regularization Network for Unsupervised 3D Brain Image Registration,'' \emph{arXiv preprint arXiv:1907.01922}, 2019.

\bibitem{b21}   P. Hu, G. Wang, and Y.-P. Tan, ``Recurrent spatial pyramid cnn for optical flow estimation,'' \emph{IEEE Transactions on Multimedia}, vol. 20, no. 10, pp. 2814-2823, 2018.

\bibitem{b22}   J.-R. Chang and Y.-S. Chen, ``Pyramid stereo matching network,'' \emph{in Proceedings of the IEEE Conference on Computer Vision and Pattern Recognition}, pp. 5410-5418, 2018.

\bibitem{b23}   D. Sun, X. Yang, M.-Y. Liu, and J. Kautz, ``PWC-Net: CNNs for optical flow using pyramid, warping, and cost volume,'' \emph{in Proceedings of the IEEE Conference on Computer Vision and Pattern Recognition}, pp. 8934-8943, 2018.

\bibitem{b24}   T.-W. Hui, X. Tang, and C. Change Loy, ``Liteflownet: A lightweight convolutional neural network for optical flow estimation,'' \emph{in Proceedings of the IEEE Conference on Computer Vision and Pattern Recognition}, pp. 8981-8989, 2018.

\bibitem{b25}   J. Hur and S. Roth, ``Iterative Residual Refinement for Joint Optical Flow and Occlusion Estimation,'' \emph{in Proceedings of the IEEE Conference on Computer Vision and Pattern Recognition}, pp. 5754-5763, 2019.

\bibitem{b26}   X. Hu, M. Kang, W. Huang, M. R. Scott, R. Wiest, and M. Reyes, ``Dual-Stream Pyramid Registration Network,'' \emph{in International Conference on Medical Image Computing and Computer-Assisted Intervention}, pp. 382-390: Springer, 2019.

\bibitem{b27}   B. D. de Vos, F. F. Berendsen, M. A. Viergever, H. Sokooti, M. Staring, and I. Išgum, ``A deep learning framework for unsupervised affine and deformable image registration,'' \emph{Medical image analysis}, vol. 52, pp. 128-143, 2019.

\bibitem{b28}   T. Lau, J. Luo, S. Zhao, E. I. Chang, and Y. Xu, ``Unsupervised 3D End-to-End Medical Image Registration with Volume Tweening Network,'' \emph{arXiv preprint arXiv:1902.05020}, 2019.

\bibitem{b29}   S. Zhao, Y. Dong, E. I. Chang, and Y. Xu, ``Recursive Cascaded Networks for Unsupervised Medical Image Registration,'' \emph{arXiv preprint arXiv:1907.12353}, 2019.

\bibitem{b30}   D. Wei, Z. Wu, G. Li, X. Cao, D. Shen, and Q. Wang, ``Deep Morphological Simplification Network (MS-Net) for Guided Registration of Brain Magnetic Resonance Images,'' \emph{arXiv preprint arXiv:1902.02342}, 2019.

\bibitem{b31}   V. Vishnevskiy, T. Gass, G. Szekely, C. Tanner, and O. Goksel, ``Isotropic total variation regularization of displacements in parametric image registration,'' \emph{IEEE transactions on medical imaging}, vol. 36, no. 2, pp. 385-395, 2016.

\bibitem{b32}   D. S. Marcus, T. H. Wang, J. Parker, J. G. Csernansky, J. C. Morris, and R. L. Buckner, ``Open Access Series of Imaging Studies (OASIS): cross-sectional MRI data in young, middle aged, nondemented, and demented older adults,'' \emph{Journal of cognitive neuroscience}, vol. 19, no. 9, pp. 1498-1507, 2007.

\bibitem{b33}   M. P. Milham, D. Fair, M. Mennes, and S. H. Mostofsky, ``The ADHD-200 consortium: a model to advance the translational potential of neuroimaging in clinical neuroscience,'' \emph{Frontiers in systems neuroscience}, vol. 6, p. 62, 2012.

\bibitem{b34}   A. Di Martino, et al, ``The autism brain imaging data exchange: towards a large-scale evaluation of the intrinsic brain architecture in autism,'' \emph{Molecular psychiatry}, vol. 19, no. 6, p. 659, 2014.

\bibitem{b35}   K. Marek, et al, ``The parkinson progression marker initiative (PPMI),'' \emph{Progress in neurobiology}, vol. 95, no. 4, pp. 629-635, 2011.

\bibitem{b36}   D. W. Shattuck, et al, ``Construction of a 3D probabilistic atlas of human cortical structures,'' \emph{Neuroimage}, vol. 39, no. 3, pp. 1064-1080, 2008.

\bibitem{b37}   A. Klein, et al, ``Evaluation of 14 nonlinear deformation algorithms applied to human brain MRI registration,'' \emph{Neuroimage}, vol. 46, no. 3, pp. 786-802, 2009.

\bibitem{b38}   J. Ashburner, ``A fast diffeomorphic image registration algorithm,'' \emph{Neuroimage}, vol. 38, no. 1, pp. 95-113, 2007.

\bibitem{b39}   T. Vercauteren, X. Pennec, A. Perchant, and N. Ayache, ``Diffeomorphic demons: Efficient non-parametric image registration,'' \emph{NeuroImage}, vol. 45, no. 1, pp. S61-S72, 2009.


\end{thebibliography}
\end{document}